\pdfoutput=1

\documentclass[11pt]{article}

\usepackage[]{acl2024}
\usepackage{multirow}
\usepackage{amssymb}
\usepackage{ragged2e}
\usepackage{times}
\usepackage{latexsym}
\usepackage{graphicx}
\usepackage{float}
\usepackage{pgfplots}
\usepackage{mathrsfs}
\usepackage{array}
\newcolumntype{P}[1]{>{\centering\arraybackslash}p{#1}}
\usepackage{supertabular}
\usepackage{longtable}
\pgfplotsset{compat=1.18, width=7.7cm}
\usepackage[T1]{fontenc}

\usepackage[utf8]{inputenc}

\usepackage{subfig}
\usepackage{overpic}   
\usepackage{float}
\usepackage{enumitem} 

\usepackage{microtype}

\usepackage{inconsolata}
\usepackage{booktabs}
\usepackage{stfloats}
\usepackage{amsmath}
\usepackage{graphicx}
\usepackage{xcolor}
\usepackage[linesnumbered,ruled,vlined]{algorithm2e}
\usepackage{algorithmic}           
\title{RA-ISF: Learning to Answer and Understand from Retrieval Augmentation via Iterative Self-Feedback}

\author{
   ~~Yanming Liu$^{1}$
   ~~Xinyue Peng$^{2}$
   ~~Xuhong Zhang$^{1}$
   ~~Weihao Liu 
   ~~Jianwei Yin$^{1}$ \\ \bf
   ~~Jiannan Cao$^{3}$
   ~~Tianyu Du$^{1}$\footnotemark[1]\\ 
   $^{1}$Zhejiang University 
   $^{2}$Southeast University \\
   $^{3}$Massachusetts Institute of Technology\\
   \texttt{{\{oceann24, zhangxuhong, zjradty\}@zju.edu.cn, zjuyjw@cs.zju.edu.cn,}}\\
\texttt{{xinyuepeng@seu.edu.cn, jiannan@mit.edu, liuweihao2022@outlook.com}}\\
}

\begin{document}

\maketitle

\renewcommand{\thefootnote}{\fnsymbol{footnote}}
\footnotetext[1]{Corresponding author.}
\renewcommand{\thefootnote}{\arabic{footnote}}

\begin{abstract}
Large language models (LLMs) demonstrate exceptional performance in numerous tasks but still heavily rely on knowledge stored in their parameters. Moreover, updating this knowledge incurs high training costs. Retrieval-augmented generation (RAG) methods address this issue by integrating external knowledge. The model can answer questions it couldn't previously by retrieving knowledge relevant to the query. This approach improves performance in certain scenarios for specific tasks. However, if irrelevant texts are retrieved, it may impair model performance. In this paper, we propose Retrieval Augmented Iterative Self-Feedback (RA-ISF), a framework that iteratively decomposes tasks and processes them in three submodules to enhance the model's problem-solving capabilities. 
Experiments show that our method outperforms existing benchmarks, performing well on models like GPT3.5, Llama2, significantly enhancing factual reasoning capabilities and reducing hallucinations.\footnote{Our code is public at \url{https://github.com/OceannTwT/ra-isf}}

\end{abstract}

\section{Introduction}

Large language models (LLMs)~\citep{brown2020language, chowdhery2023palm, touvron2023Llama} have demonstrated their excellent performance in knowledge reasoning and outstanding capabilities across various task domain~\citep{bang2023multitask, ouyang2022training}. However, the parameterized knowledge stored within LLMs may be incomplete and hard to incorporate up-to-date knowledge~\citep{dhingra-etal-2022-time,10.1145/3383123}. To address this issue, retrieval-augmented generation (RAG) approaches can leverage external knowledge and documents, extract non-parameterized knowledge, and incorporate it into the model's prompts, thereby embedding new knowledge into the language model~\citep{ram2023context, guu2020retrieval}. This approach demonstrates outstanding performance in answering a variety of open-domain questions.

However, current RAG frameworks have two major challenges. First, retrieving irrelevant knowledge texts will impair the LLMs' ability to solve tasks~\citep{shi2023large, mallen2023not}. Second, the incorporation of LLM's existing knowledge and the retrieved knowledge may face difficulty~\citep{izacard2022atlas}. Some methods have conducted research based on these issues, including considering the model's problem-solving abilities~\citep{wang-etal-2023-self-knowledge} and whether the retrieved passages are relevant to the question~\citep{chen2023understanding, asai2024selfrag, yu2023chainofnote}. However, current solutions still have drawbacks in answering knowledge-intensive questions and different levels of sub-questions. Therefore, how to fuse knowledge and utilize knowledge for question answering is very important in this process.

To overcome the above limitations, we introduce \textbf{R}etrieval \textbf{A}ugmented \textbf{I}terative \textbf{S}elf-\textbf{F}eedback (\textbf{RA-ISF}), a framework addresses problems by iteratively processing questions.
Specifically, unlike directly appending retrieved knowledge into prompts, our approach employs three sub-modules for iterative processing. These three sub-modules are the \textit{Self-Knowledge Module}, the \textit{Passage Relevance Module}, and the \textit{Question Decomposition Module}. We have also collected a series of data through LLMs to evaluate whether a specific module possesses the corresponding capabilities. By training a small language model or simply relying on in-context learning, these modules can demonstrate capabilities in self-knowledge, relevance judgment, and question decomposition.

\begin{figure*}[tp]
    \centering
    \includegraphics[width=1\linewidth]{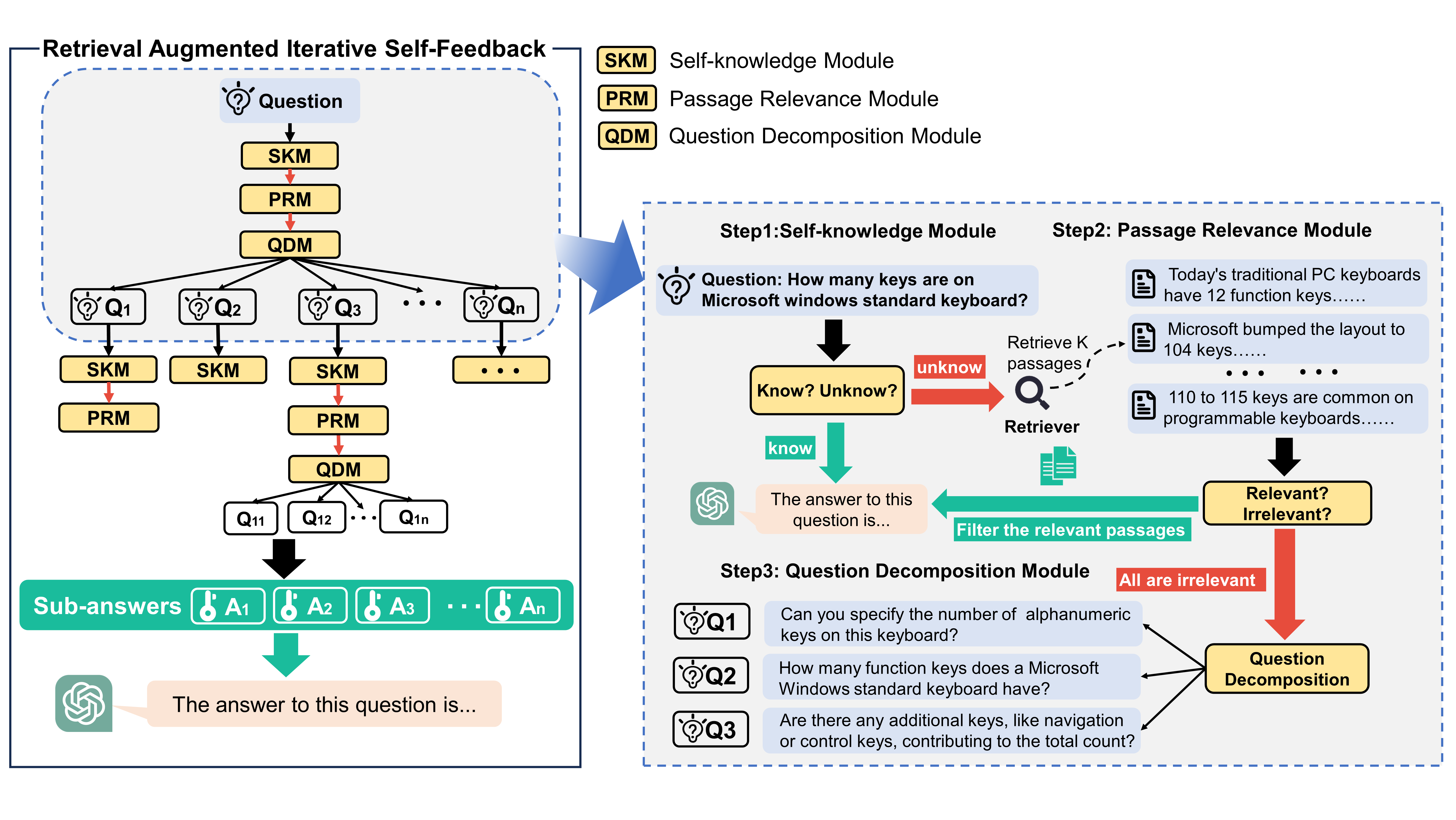}
    \caption{Overview of RA-ISF. It consists of three sub-modules: self-knowledge, passage relevance, and question decomposition. 
    }
    \label{fig:overview}
\end{figure*}

As shown in Figure \ref{fig:overview}, RA-ISF first uses a self-knowledge module to determine whether the current question could be answered on its own knowledge. Then, when employing a retrieval strategy, the passage relevance module will assess the relevance of each retrieved paragraph to the problem. Relevant paragraphs will be integrated into the prompt and used for prediction. When all paragraphs are irrelevant to the question, the question decomposition module will break down the questions into sub-questions and repeat the aforementioned steps for these sub-questions. Ultimately, the model will synthesize the answers to the sub-questions to respond to the original question.

Compared to previous RAG methods, our iterative self-feedback approach more effectively unleashes the potential of the model and better incorporates external knowledge with the model's inherent knowledge. Simultaneously, RA-ISF can address questions by decomposing them when the model lacks an initial answer or retrieves irrelevant texts, combining these solutions to answer the origin question, which is an effective problem-solving strategy. Experiments on various LLMs (e.g., GPT3.5~\citep{openai2023gpt} and Llama-2~\citep{touvron2023Llama}) demonstrate that RA-ISF exhibits superior performance in handling complex questions compared to existing methods.

\textbf{Our Contributions.} Our main contributions are summarized as follows. 
\begin{itemize}
\item[$\bullet$]  We introduce RA-ISF, an innovative retrieval-augmented framework designed to tackle diverse challenges. This approach evaluates the model's ability to solve the corresponding problem and its relevance to the retrieved content through an iterative method. This comprehensive evaluation is crucial for solving complex problems.
\item[$\bullet$] 
To the best of our knowledge, this is the first time an iterative question decomposition approach has been used in a retrieval-augmented framework, which mitigates the impact of irrelevant text interference.
\item[$\bullet$] Our proposed framework significantly enhances knowledge retrieval performance across different tasks, demonstrating the potential and robustness of our framework. 

\end{itemize}

\section{Related Work}
\subsection{Retrieval Augmented Language Model}
The retrieval-augmented language model (LM) is enhanced by a non-parametric memory
to facilitate external knowledge access and provide provenance~\citep{guu2020retrieval, lewis2020retrieval, shi2023replug}. However, the improved task performance of retrieval augmentation largely depends on the relevance of the retrieved passage~\citep{shi2023large}. Recently, some studies have begun to explore when to use retrieval for diverse instruction. For instance, \citet{asai2024selfrag} integrates special feedback tokens into the language model to the need for retrieval and confirm the output's relevance, support, or completeness. \citet{chen2023understanding} investigates the impact of texts with different attributes and relevance on text generation performance.
Some works~\citep{mallen2023not} explore the incorporation of the LLM's inherent knowledge with in-context documents. \citet{wang2023self} improves the performance in answering self-knowledge questions by guiding the model to acquire self-knowledge capabilities. Meanwhile, other studies have concentrated on iterative retrieval augmentation \citep{trivedi-etal-2023-interleaving, shao-etal-2023-enhancing} and accelerating retrieval speed \citep{xu2023recomp}. 

In comparison, our method combines the model's retrieval and understanding capabilities and reduces its susceptibility to irrelevant texts. This is achieved through the task decomposition paradigm. By iteratively processing these three sub-modules with self-feedback, we develop a versatile and robust retrieval-augmented framework.

\subsection{Task Decomposition}
Task decomposition is an effective method for solving knowledge-intensive and other complex tasks. It involves breaking down multi-turn questions into single-turn questions, answering each sub-task separately, and then synthesizing these answers to resolve the original task. \citet{perez2020unsupervised} trains a question decomposition and task aggregation model to split and collectively solve the original problem. \citet{yang2022seqzero} decomposes questions into a series of slot-filling tasks, transforming natural language questions into SQL queries, and implements natural language prompts corresponding to SQL clauses through a rule-based system. Least-to-most~\citep{zhou2023leasttomost} leverages the in-context learning capabilities of large language models, solving problems by providing examples of question decomposition.

RA-ISF utilizes task decomposition to mitigate the impact of irrelevant prompt texts on the model~\citep{shi2023large}, by iteratively answering sub-questions and integrating text relevance with self-knowledge answering capabilities into the framework. This enhances the performance in solving the entire problem.

\section{Methodology}
Existing retrieval augmented methods still have some shortcomings. For instance, the model may struggle to solve problems based solely on its own knowledge, and during retrieval, it might be influenced by irrelevant texts, leading to the generation of incorrect answers. Therefore, we introduce an upgraded retrieval-augmented generation 
framework -- Retrieval Augmented Iterative Self-Feedback (RA-ISF), which improves the quality and accuracy of LLM responses through internal knowledge comprehension, external knowledge retrieval, and problem decomposition.

\subsection{Overview }
As shown in Figure \ref{fig:overview}, RA-ISF involves three pre-trained models: $\mathcal{M}_{know}$, $\mathcal{M}_{rel}$ and $\mathcal{M}_{decom}$, each responsible for internal knowledge assessment, external knowledge retrieval, and question decomposition functions, respectively.

In general, we input a question $q_{new}$ and obtain its answer $\mathcal{A}$ through the RA-ISF framework. The overall process is as follows: Firstly, input $q_{new}$ into $\mathcal{M}_{know}$ to determine if it can be solved using internal knowledge. If solvable, directly output the answer. If not, use the retriever $\mathcal{R}$ to search for relevant information for the question $q_{new}$. Combine the retrieved text with the question and input them into the model $\mathcal{M}_{rel}$ to assess their relevance. If relevant, generate an answer based on these related passages. If none of the retrieved text is relevant, input $q_{new}$ into the question decomposition model $\mathcal{M}_{decom}$ to break it down into multiple sub-questions $q_{1}$, ..., $q_{n}$. Next, input these sub-questions back into the model $\mathcal{M}_{know}$ (and $\mathcal{M}_{rel}$, $\mathcal{M}_{decom}$ if needed) to obtain corresponding sub-answers. Finally, integrate these sub-answers to generate the ultimate answer.

\subsection{RA-ISF Training }
In this section, we will delve into the training process of the models within RA-ISF, encompassing both dataset collection and model learning. Due to the similarity in the training procedures for the three models, we will use the training of the $\mathcal{M}_{know}$ model as an illustrative example.

\vspace{5pt}

{
\setlength{\parindent}{0cm}

\textbf{Data Collection.} First, we need to construct a dataset generated by LLMs. Specifically, based on various training objectives, we collect corresponding questions $\mathcal{Q}=\{Q_1, Q_2, ..., Q_n\}$ and input them one by one into the LLM model $\mathcal{M}$. By providing the model with specific instructions to perform the respective tasks, and utilizing few-shot prompts and in-context learning, we enable model $\mathcal{M}$ to generate answers $\mathcal{A}=\{A_1, A_2, ..., A_n\}$ corresponding to each question.
}

 We have collected various types of supervised training data, and through the previously described process, combined them into the training data for the model. Ultimately, this resulted in a trained dataset $\mathcal{D^*}=\{\mathcal{Q},\mathcal{A}\}$. For specific details on the data collection process for each sub-model $\mathcal{M}_{know}$, $\mathcal{M}_{rel}$, $\mathcal{M}_{decom}$, please refer to Appendix \ref{sec:appendix A}.

\vspace{5pt}

{
\setlength{\parindent}{0cm}
\textbf{Model Learning.} 
After collecting the training data $\mathcal{D}^{*}$, we initialize $\mathcal{M}_{sub}$ using a pre-trained language model and train it on $\mathcal{D}^{*}$ using a standard conditional language modeling objective to maximize the effectiveness of classification. Here, we use cross-entropy loss to represent this, denoted as:}
\begin{equation}
\min_{\mathcal{M}_{sub}} -\mathbb{E}_{(\mathcal{Q},\mathcal{A})\sim \mathcal{D}^{*} }\log P_{\mathcal{M}_{sub}}(\mathcal{A}\mid \mathcal{Q}).
\end{equation}
The initial model can be any pre-trained language model. Here, we initialize $\mathcal{M}_{sub}$ using the Llama 2-7B model \citep{touvron2023Llama}.

\subsection{RA-ISF Inference}

\title{Algorithm template}
\begin{algorithm}[h]
\caption{Problem Iterative Solving}\label{algorithm1}
\SetKwInput{KwInput}{Input}                
\SetKwInput{KwOutput}{Output}              
\DontPrintSemicolon
  \KwInput{$q_{new}$, $\mathcal{M}_{know}$, $\mathcal{M}_{rel}$, $\mathcal{M}_{decom}$, $\mathcal{M}$, $\mathcal{R}$, $\mathcal{C}$}
  \KwOutput{$\mathcal{A}$}
  \SetKwFunction{FMain}{Main}
  \SetKwFunction{FSum}{Sum}
  \SetKwFunction{FPB}{Problem-solving}
    
  \SetKwProg{Fn}{Function}{:}{}
  \Fn{\FPB{$q_{t}$, iter}}{
  \If{iter > $D_{th}$}
  {
  $\mathcal{A}$ = $\textit{Unknow}$\;
  \KwRet $\mathcal{A}$\;}
    \If{$\mathcal{M}_{know}(q_{t})$=Know}
    {
     $\mathcal{A} \leftarrow\mathcal{M}(q_{t})$\;
     \KwRet $\mathcal{A}$\;
    }
     $\mathcal{P}=\left \{p_1, p_2, …, p_k  \right \}  \leftarrow \mathcal{R} (q_{t},\mathcal{C})$\;
    $\mathcal{P}_{rel}=\varnothing$\;
    \For{$i=1$ \KwTo $k$}{
    \If{ $\mathcal{M}_{rel}(p_i)$=relevant }{
    $\mathcal{P}_{rel}= \mathcal{P}_{rel}\cup {p_i}$\;
    }
    }
    \If{size$(\mathcal{P}_{rel})>0$}
    {
         $\mathcal{A} \leftarrow  \mathcal{M}(q_{t},\mathcal{P})$\;
         \KwRet $\mathcal{A}$\;
    }
     ${Q}_{sub}=\left\{ q_1, ..., q_n  \right \}  \leftarrow \mathcal{M}_{decom}( q_{t})$\;
    \For{$i=1$ \KwTo $n$ }{
    $a_i$ =\texttt{Problem-solving}($q_i,iter+1)$\;
    $\mathcal{A}_{sub}$ = $a_i\cup\mathcal{A}_{sub}$  \;
    } 
     $\mathcal{A} \leftarrow  \mathcal{M}(q_{t}, {Q}_{sub}, \mathcal{A}_{sub})$\;
    \KwRet $\mathcal{A}$\;
  }
  $\mathcal{A} \leftarrow \texttt{Problem-solving}(q_{new},0)$\;
\end{algorithm}

In this section, we provide a detailed explanation of how the RA-ISF framework infers and predicts answers for the question $q_{new}$. Algorithm \ref{algorithm1} presents the details of RA-ISF at inference. 
Note that we use three pre-trained models $\mathcal{M}_{know}$, $\mathcal{M}_{rel}$, $\mathcal{M}_{decom}$, the LLM for answering questions  $\mathcal{M}$, the retriever $\mathcal{R}$, and the corpus $\mathcal{C}$. Additionally, we have the question $q_{new}$ to be addressed.

\vspace{5pt}

{
\setlength{\parindent}{0cm}

\textbf{Self-Knowledge Inference.} 
The RA-ISF framework utilizes the $\mathcal{M}_{know}$ model to infer whether the question $q_{new}$ can be addressed using the model's own knowledge. If so, the question is input into $\mathcal{M}$ to directly predict the answer $\mathcal{A}$. The formal expression is as follows:
}
\begin{equation}
\mathcal{A}= {\mathop{\arg\max}\limits_{a}} \ P(a|q_{new}).
\end{equation}
If $\mathcal{M}$ cannot use its own knowledge to solve the question $q_{new}$, we move to the next step.

\vspace{5pt}
{
\setlength{\parindent}{0cm}

\textbf{Passages Relevance Inference.} 
 When the model cannot solve the question $q_{new}$ using its internal knowledge, we use the retriever $\mathcal{R}$ to search for the most suitable $k$ passages $\mathcal{P}=\left \{p_1, p_2, …, p_k  \right \} $ in the corpus $\mathcal{C}$.
}
Since the retriever may find passages unrelated to the question, potentially leading to erroneous answers, we need to filter the retrieved passages. Here, we use ``relevance'' as the criteria, evaluated by the model $\mathcal{M}_{rel}$.

Suppose $n (n=0, 1, …, k$) relevant passages $\mathcal{P}_{rel}$  are finally filtered. If $n > 0$, these $n$ passages are used as prompts, combined with $q_{new}$, and input into the model $\mathcal{M}$ to obtain the final answer $\mathcal{A}$. The formal expression is as follows:
\begin{equation}
\mathcal{A} = {\mathop{\arg\max}\limits_{a}} \ {P}(a|q_{new},\mathcal{P}_{rel}).
\end{equation}
If $n=0$, which means all the retrieved passages are irrelevant to the question, we proceed to the next step. 

\vspace{5pt}

{
\setlength{\parindent}{0cm}

\textbf{Problem Decomposition.} 
If $q_{new}$ cannot be solved using its own and external knowledge, we will decompose complex questions into a series of simpler sub-problems for resolution.

}
In this process, we employ the $\mathcal{M}_{decom}$ model to decompose $q_{new}$ into multiple sub-problems ${Q}_{sub}=\left\{ q_1, ..., q_n  \right \} $. Subsequently, we take each sub-problem reintroduce it to the $\mathcal{M}_{know}$ model (determining the use of $\mathcal{M}_{rel}$ and $\mathcal{M}_{decom}$ based on the specific condition), and obtain corresponding sub-answers $\mathcal{A}_{sub}$.
If a sub-problem $q_k$ has been iteratively decomposed $D_{th}$ times, we consider that the model cannot find the answer to this problem, and the answer for $a_k$ is set as ``unknown''. 

Once we have the answers $\mathcal{A}_{sub}= \left \{ a_1, ..., a_n \right \} $ for all sub-problems, we use all the sub-problems ${Q}_{sub}$ and their answers $\mathcal{A}_{sub}$ as prompts for $q_{new}$. Then input them all into the model $\mathcal{M}$ to predict the answer $\mathcal{A}$ for this question. The formal expression is as follows:
\begin{equation}
\mathcal{A} = {\mathop{\arg\max}\limits_{a}} \  {P}(a|q_{new},\mathcal{A}_{sub},{Q}_{sub}).
\end{equation}

\section{Experimental Setup}

\subsection{Datasets}

To comprehensively evaluate performance in datasets with different characteristics, we use the following five representative datasets for evaluation: Natural Question (NQ)~\citep{kwiatkowski2019natural}, TriviaQA~\citep{joshi2017triviaqa}, StrategyQA~\citep{geva2021did}, HotpotQA~\citep{yang2018hotpotqa}, and 2WikiMQA~\citep{ho2020constructing}.

\subsection{Models}

The models in our framework fall into two categories: an LLM for prediction and three models that serve as intermediate steps to assess the problem's characteristics. For the LLM, we experiment with open-sourced Llama2~\citep{touvron2023Llama} of various sizes, as well as the GPT-3.5 (text-davinci-003)~\citep{openai2023gpt} through the OpenAI API. As for the three sub-models, we employ Llama2-7b as their pre-trained model. 

\subsection{Retriever and Corpus}
For fair evaluation, we use the same retriever for different approaches to search the same corpus. Specifically, we employ Contriever-MS-MARCO~\citep{izacard2022unsupervised} 
as the retriever and use the corpus from Wikipedia as of Dec. 20, 2018~\citep{karpukhin2020dense}. These articles are segmented into non-overlapping fragments of 100 words. To avoid contamination, we remove input prompts $x$ from the corpus that is contained in the dataset. To prevent the dilution of useful information, we follow \citet{ram2023context} and set the retrieval length to $l$ = 64.

\begin{table*}[ht!]
\tiny
    \renewcommand{\arraystretch}{1}
	 \resizebox{\linewidth}{!}{
\begin{tabular}{lcccccc}
\toprule
\textbf{Method} & \textbf{Avg.}          & \textbf{NQ}            & \textbf{TriviaQA}      & \textbf{HotpotQA}      & \textbf{StrategyQA}    & \textbf{2WikiMHQA}     \\ \midrule

\multicolumn{7}{c}{\textbf{\textit{GPT3.5 Without Retrieval}}}                         \\ \midrule
 Direct                  & 41.8          & 29.2          & 67.3          & 22.1          & 65.2          & 23.6         \\
 Least-to-most                 &  46.3         & 32.5         & 68.8          & 30.2          & 68.5          & 31.3          \\ \midrule

\multicolumn{7}{c}{\textbf{\textit{GPT3.5 With Retrieval}}}    \\ \midrule
 IRCoT                & 46.5          &  32.9         & 66.8          & 33.7          & 67.9         & 31.1          \\
 RAG                 & 44.2          & 31.7          & 64.2          & 32.2          & 64.7          & 28.4          \\
 $\text{SKR}_\text{knn}$                &  47.6         & 33.8          &  67.5         &  34.2         & 70.1          &  32.5         \\
 $\text{Iter-RetGen}_\text{3}$                &  -         & -          &  -         &  45.2*         & 72.3*          &  34.8*         \\
 RA-ISF(ours)                & \textbf{55.0}          & \textbf{40.2}          & \textbf{76.1}          & \textbf{46.5}         & \textbf{75.9}          & \textbf{36.1}        
  \\ \midrule

\multicolumn{7}{c}{\textbf{\textit{$\text{Llama-2}_\text{13b}\text{ Without Retrieval}$}}}                         \\ \midrule
  Vanilla LM                    & 27.1          & 17.4          & 38.5          & 14.0          & 52.2          &13.3          \\
 Least-to-most                 & 32.9          & 22.8         & 45.2          & 15.8          & 60.5          & 20.1          \\\midrule

\multicolumn{7}{c}{\textbf{\textit{$\text{Llama-2}_\text{13b}\text{ With Retrieval}$}}}                         \\ \midrule
 IRCoT                & 34.0          & 23.4          & 48.3          & 17.1          & 59.1          & 21.9          \\
 RAG                 & 33.9          & 21.6          & 47.0          & 17.6          & 60.8          & 22.4          \\
 $\text{SKR}_\text{knn}$                & 36.0          & 20.8          & 55.4          & 18.9          & 61.6          & 23.2  \\
 \text{REPLUG}                &  38.6         & 23.8          &  58.6         &  21.8         & 62.9          &  25.7         \\
 $\text{Self-RAG}_\text{13B}$               & 44.1         & 28.4          & 69.3          & 25.4          & \textbf{67.2}          & 30.2    \\
 RA-ISF(ours)                & \textbf{46.0}         & 
 \textbf{31.3}          & \textbf{71.4}          & \textbf{28.9}          & 66.7          & \textbf{31.7}    
\\ \bottomrule
\end{tabular} }
    \caption{Main experimental results. 
    \textbf{Bold} number indicates the best performance among all methods in this model. * indicates the results from the original paper.}
    \vspace{-0.15cm}
\label{tab:main_result}
\end{table*}

\subsection{Baselines}

To conduct a holistic evaluation and comparison, 
 we use the same datasets, with the same retriever and corpus to compare our method with the following baselines:

\vspace{5pt}

{
\setlength{\parindent}{0cm}
\textbf{Directly Prompting and Vanilla LM}~\citep{brown2020language}  involves presenting questions directly to the LLM, prompting it to generate corresponding answers without any explanations. 
}

\vspace{5pt}

{
\setlength{\parindent}{0cm}
\textbf{Least-to-most}~\citep{zhou2023leasttomost} guides the LLM to break down the question and assist in solving the original problem by answering sub-questions.
}
\vspace{5pt}

{
\setlength{\parindent}{0cm}
\textbf{IRCoT}~\citep{trivedi-etal-2023-interleaving} enhances each step of the chain-of-thought generation process by incorporating knowledge retrieval steps during the generation process.
}

\vspace{5pt}

{
\setlength{\parindent}{0cm}
\textbf{RAG}~\citep{guu2020retrieval,lewis2020retrieval} assists in answering questions by retrieving information from external documents. We append the retrieved passage to the question in the experiment.
}

\vspace{5pt}

{
\setlength{\parindent}{0cm}
\textbf{SKR}~\citep{wang-etal-2023-self-knowledge} trains a small model to determine whether the LLM can answer a question using its own knowledge, and decides whether to perform retrieval for the given question.
}

\vspace{5pt}

{
\setlength{\parindent}{0cm}
\textbf{REPLUG}~\citep{shi2023replug} adapts the framework to the corresponding downstream tasks by fine-tuning the Retriever. This method enhances retrieval effectiveness by improving the relevance of the retrieved text.
}

\vspace{5pt}

{
\setlength{\parindent}{0cm}
\textbf{Iter-RetGen}~\citep{shao-etal-2023-enhancing} conducts retrieval based on multiple iterations, relying on the content retrieved in each round to aid in finding more text information relevant to the question.
}

\vspace{5pt}

{
\setlength{\parindent}{0cm}
\textbf{Self-RAG}~\citep{asai2024selfrag} provides a framework by training a LLM to learn specific reflection tokens, thereby controlling the decision of whether to retrieve during reasoning and examining the relevance of the retrieved content. We compares our method with the open-source $\text{Self-RAG}_\text{13b}$.

}

\subsection{Implementation Details}
We randomly sampled 1000 input prompts from each dataset and generated labels or answers (Relevance, Self-Knowledge) for these prompts using GPT-4. The labels or answers are then used to fine-tune these three pre-trained models. For these three models, we adopt a learning rate of 5e-4 during training. Greedy decoding is consistently used in the inference process across all experiments to maintain deterministic generation outcomes. This distillation process allows us to augment the pre-trained model with feature analysis capabilities. The default iteration threshold is set to 3.
To evaluate the effectiveness of the method, we use \textbf{Exact Match} as our standard metrics.









\section{Experiment Results}

\subsection{Main results}

The main results are shown in Table~\ref{tab:main_result}. From the results, we have the following observations.

\vspace{5pt}

{
\setlength{\parindent}{0cm}

\textbf{Our proposed RA-ISF outperformed other methods on all five datasets on GPT3.5.} On average, the performance improvement of RA-ISF is \textbf{+8.7} compared to the baseline without retrieval. Compared to the baseline with retrieval, RA-ISF surpasses all existing methods, achieving an average performance improvement of \textbf{+7.4} compared to the optimal method. In addition, compared to Iter-RetGen, which also uses iterative retrieval, RA-ISF shows an improvement of \textbf{+2.0} on HotpotQA, StrategyQA, and 2WikiMHQA.

}

\vspace{5pt}

{
\setlength{\parindent}{0cm}

\textbf{RA-ISF is also effective on smaller-scale LLMs.} We experimented with our approach on $\text{Llama2}_\text{13B}$, and the results showed that our method achieved SOTA on four out of five datasets, with an average improvement of +1.9 compared to the best-performing $\text{Self-RAG}_\text{13B}$. The performance of $\text{Llama2}_\text{13B}$ on multiple datasets reaches or even surpasses GPT-3.5 + RAG, highlighting the assistance of our method in problem-solving.

}

\vspace{5pt}

{
\setlength{\parindent}{0cm}

\textbf{RA-ISF helps alleviate the hallucination problem associated with RAG.} For instance, in TriviaQA and StrategyQA datasets, Direct RAG leads to a decrease in performance, possibly due to the negative impact of irrelevant retrieval content~\citep{shi2023large}. In our framework, three sub-modules help the model to reduce hallucinations and enhance knowledge representation. Compared to GPT-3.5 + RAG, our GPT-3.5 + RA-ISF achieves a \textbf{+11.2} performance improvement on StrategyQA. Similar performance improvements are observed on TriviaQA as well.

}

\subsection{Ablation Studies}

\begin{table}[]
	 \resizebox{\linewidth}{!}{
\begin{tabular}{lccc}
\toprule
\multicolumn{1}{c}{\textbf{Method}} & \begin{tabular}[c]{@{}c@{}}\textbf{NQ}\\ (\textbf{EM})\end{tabular}   & \begin{tabular}[c]{@{}c@{}}\textbf{TriviaQA}\\ (\textbf{EM})\end{tabular} & \begin{tabular}[c]{@{}c@{}}\textbf{HotpotQA}\\ (\textbf{EM})\end{tabular} \\ \midrule
Direct  & 29.2 & 67.3     & 22.1     \\ \midrule
RAG                  & 31.7 & 64.2     & 32.2     \\
Least-to-Most        & 32.5 & 68.8     & 30.2     \\
RA-ISF               & \textbf{40.2} & \textbf{76.1}     & \textbf{46.5}     \\ \midrule
\textit{No SKM}         & 37.9 & 72.3     & 40.1    \\
\textit{No PRM}         & 35.8 & 70.3    & 34.7     \\
\textit{No QDM}         & 34.6 & 71.5     & 34.9     \\ \bottomrule
\end{tabular}}
\caption{Ablation of different components on GPT3.5. \textit{No SKM}, \textit{No PRM}, and \textit{No QDM} stand for removing the submodel of Self-Knowledge, Passage-Relevant, and Question Decomposition.}
\label{tab:ablation}
\end{table}

To assess the impact of different components of RA-ISF, we set up three variants:
\begin{itemize}[leftmargin=1em]

\item[$\bullet$]
{
\setlength{\parindent}{0cm}
\textit{No Self-Knowledge Module}: This variant processes questions directly through the Passage Relevant Module without self-knowledge judgment.
}
\item[$\bullet$]
{
\setlength{\parindent}{0cm}
\textit{No Passage-Relevant Module}: After self-knowledge judgment, if the Self-Knowledge Module indicates the answer can not be addressed using the model's own knowledge, it directly decomposes the question without involving the Passage-Relevant module.
}
\item[$\bullet$]
{
\setlength{\parindent}{0cm}
\textit{No Question Decomposition Module}: After assessing passage relevance through the Passage-Relevant module, if no relevant paragraphs are found, the answer is marked as "unknown," and the Question Decomposition Module does not iterate. This means the RA-ISF iteration count is set to 0.
}
\end{itemize}

We conducted tests on NQ, TriviaQA, and HotpotQA datasets, comparing the results with RAG, RA-ISF, and LTM methods. All experiments use GPT3.5 as the base model.

\vspace{5pt}

{

\textbf{All three submodules contribute to better problem-solving performance.} Table~\ref{tab:ablation} presents the ablation experiment results, indicating that removing any component of RA-ISF leads to a performance decline. This suggests the importance of each component in the framework. Compared to RAG, the \textit{No Self-Knowledge Module} variant achieves better performance by decomposing unrelated text, resulting in improved results. In contrast to the Least-to-Most prompting method, \textit{No Self-Knowledge Module} variant achieves higher accuracy by prompting the language model with retrieved paragraphs (+6.3 on Average). When comparing Least-to-Most with variant \textit{No Passage-Relevant Modules}, the latter first assesses self-knowledge and then iteratively decomposes information. This variant outperforms the traditional Least-to-Most paradigm. Therefore, the iterative combination of these three components not only enhances the effectiveness of RAG but also addresses certain issues (e.g., hallucinations) after retrieval and mitigates negative impacts caused by irrelevant retrieved paragraphs.
}

\subsection{Iterations in Problem Decomposition}

\begin{figure}[tp]
    \centering
    \includegraphics[width=1\linewidth]{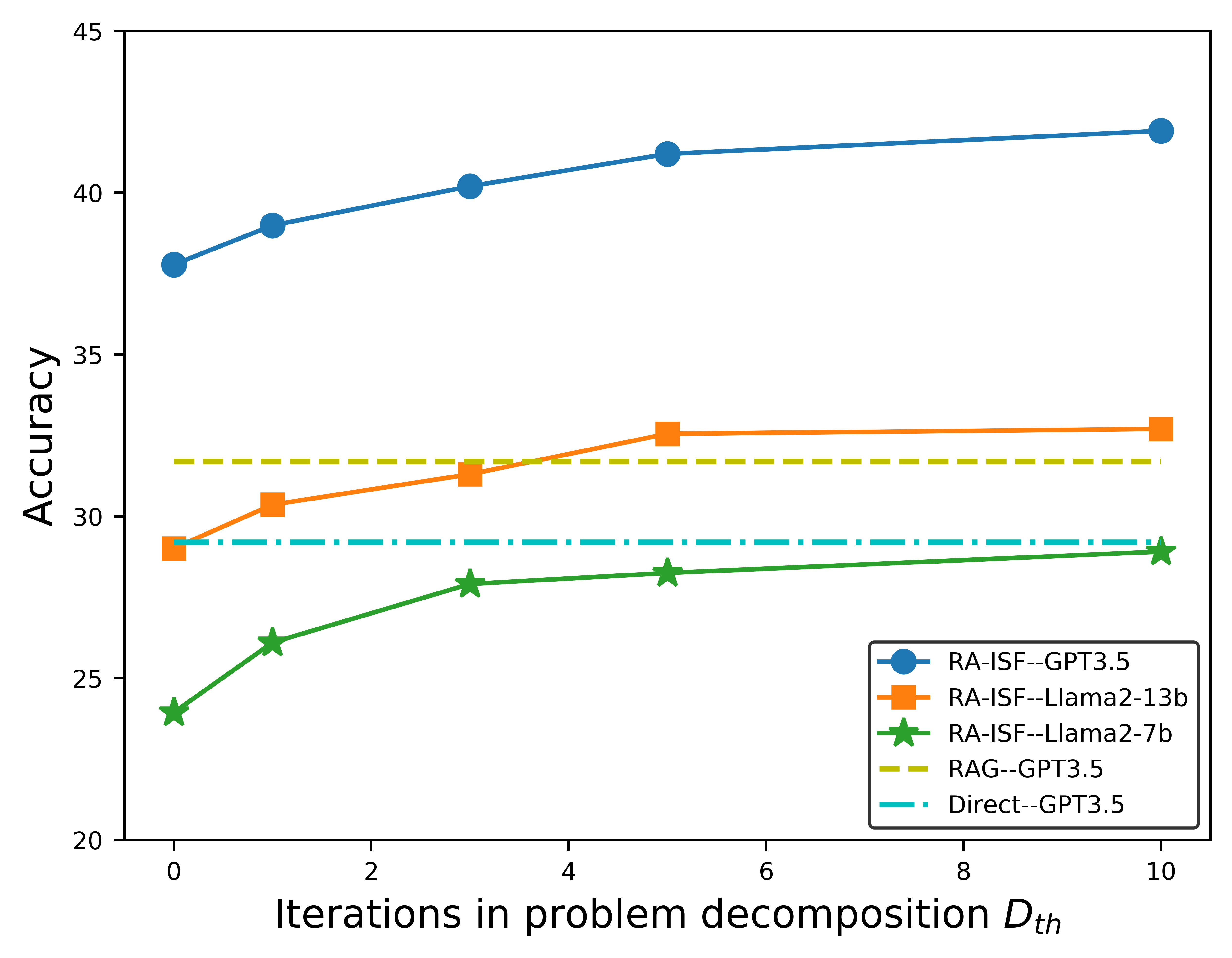}
    \caption{Question accuracy on the NQ dataset with the growth of the iteration in question decomposition $D_{th}$.}
    \label{fig:enter-label}
\end{figure}

RA-ISF sets a threshold $D_{th}$ to limit the iteration times of problem decomposition. Here, we experiment with different values of $D_{th}$ on the NQ dataset of GPT-3.5 and $\text{Llama2}_\text{7B,13B}$. Additionally, we compare RAG and Direct Prompting with RA-ISF on GPT-3.5. The accuracy of problem-solving varies with changes in $D_{th}$ as shown in Figure \ref{fig:enter-label}.

\vspace{5pt}

{

\textbf{More iterations contribute to improved performance.}
The results indicate that as the value of $D_{th}$ increases, the model's accuracy in answering questions improves. With the increase of $D_{th}$, the performance gap between RA-ISF + GPT3.5 and RAG + GPT3.5 gradually rises. More iterations also help improve the performance of small-scale LLMs in problem-solving. With the increase of $D_{th}$, the performance of RA-ISF + $\text{Llama2}_\text{13B}$ surpasses the performance of RAG and Direct Prompting on GPT3.5, and the performance on $\text{Llama2}_\text{7B}$ gradually approaches the accuracy of Direct on GPT3.5. This indicates that the iterative decomposition of problems contributes to enhancing the model's problem-solving ability.

}

\vspace{5pt}

{

\textbf{Problem decomposition helps LLM to understand.}
The goal of problem decomposition is to address situations where the model has on-parametric knowledge but struggles to answer due to inadequate understanding of the question. When $D_{th}$ is relatively small, decomposing the problem helps the model extend its problem-solving approach through reasoning and derive answers. When iteration becomes larger, it indicates that after multiple rounds of knowledge retrieval and problem decomposition, no relevant passage or on-parametric knowledge has been found. This implies that the inability of the model to solve the problem is actually due to a lack of knowledge rather than insufficient understanding. At this point, further problem decomposition is less likely to be beneficial and may even introduce misleading factors, such as decomposing unrelated sub-problems to the original question, potentially reducing the accuracy of the answers.

}
\subsection{Small Sub-model Alternatives}
In this paper, we choose the $\text{Llama2}_\text{7B}$ model as the pretrain model when training three sub-models. Since Llama2 is a 7B LM, we also want to explore the effectiveness of using a smaller model as an intermediate component. We select the T5$_\text{780M}$ model for training and compare it with Llama2$_\text{7B}$, while the base model is GPT3.5. The accuracy comparison is shown in Table~\ref{tab:submodel_size}.
\begin{table}[h]
\resizebox{\linewidth}{!}{%
\begin{tabular}
{@{}lccccc@{}}
\toprule
 & \textbf{NQ} & \textbf{TriviaQA} & \textbf{HotpotQA} & \textbf{StrategyQA} & \textbf{2WikiMHQA} \\ \midrule
$\text{Llama2}_\text{7B}$ & 40.2 & 76.1 & 46.5 & 75.9 & 36.1 \\
$\text{T5}_\text{780M}$ & 39.6 & 74.8 & 45.8 & 74.7 & 35.3 \\
GPT3.5 & 42.3 & 76.8 & 47.7 & 76.5 & 36.9 \\ \bottomrule
\end{tabular} }
\caption{Evaluation for different sizes of sub-model in various datasets.}
\label{tab:submodel_size}
\end{table}

\textbf{Training RA-ISF with a small model also yields excellent performance.} When the RA-ISF method is trained on the small T5$_\text{780M}$  model,  the accuracy of answering questions using this model is only slightly lower by one to two percentage points compared to $\text{Llama2}_\text{7B}$. This indicates that when training the three sub-models of RA-ISF, if there are constraints or cost limitations, using a small model like T5$_\text{780M}$  as the pre-trained model can still demonstrate excellent performance.

\subsection{Efficiency Analysis}

Considering that the main challenge with naive iterative problem decomposition lies in the integration with retrieval augmented, if we assume decomposing $k$ problems each time for $i$ iterations, then we would need to conduct $\frac{k^{i+1}-1}{k-1}$ times of retrievals. This approach is highly inefficient due to the substantial amount of problem merging and retrieval involved. Therefore, the Passage Relevance Module serves as a checkpoint in our structure. When relevant passage is discovered during the retrieval process for a sub-problem, we return the answer from that point. The Self-Knowledge Module serves a similar purpose, but it reduces the number of calls to the retriever. The result of retrievals and decomposition is shown on Table~\ref{tab:efficiency}.

\begin{table}[]
\resizebox{\linewidth}{!}{%
\begin{tabular}{lcccccc}
\toprule
\multicolumn{1}{c}{\multirow{2}{*}{}} & \multicolumn{2}{c}{\textbf{NQ}} & \multicolumn{2}{c}{\textbf{TriviaQA}} & \multicolumn{2}{c}{\textbf{StrategyQA}}  \\ \cmidrule(lr){2-3} \cmidrule(lr){4-5} \cmidrule(lr){6-7} 
\multicolumn{1}{c}{}                  & Ret         & Sub      & Ret            & Sub         & Ret            & Sub                   \\ \midrule
$\text{Llama2}_\text{13B}$                       & 3.4       & 5.3      & 2.9          & 4.0         & 3.6          & 6.8                  \\
$\text{GPT3.5}$                      & 2.8       & 4.5      & 2.6          & 4.1         & 3.2         & 5.1           \\
\bottomrule
\end{tabular}
}
\caption{Evaluation of average number of retrievals(Ret) and subproblems(Sub) in one query.}
\label{tab:efficiency}
\end{table}

\textbf{RA-ISF enables the model to focus more on addressing difficult subproblems while quickly resolving the simpler parts. }Through such processing, with a constraint of three iterations, the average number of retrievals and the number of question and subquestions in one query required in our experiments can be optimized to an average of 3.2 and 5.1 respectively (using the StrategyQA dataset as an example). In other words, most simple or retrievable problems are handled in the initial layers, allowing our structure to focus on potentially complex and challenging sub-problems. RA-ISF organically combines the paradigms of iterative and recursive retrieval.

\subsection{Human and Model Assessments}
\label{sec:human}
We conduct both manual and automated assessments to evaluate the reliability of RA-ISF. Specifically, we randomly select 40 questions from each dataset and invite 50 human annotators to assess the precision of the generated responses compared to GPT-4. For $\mathcal{M}_{know}$, if the model's judgment on whether the question can be answered using its own knowledge is consistent with GPT-4, it is considered precise. For $\mathcal{M}_{rel}$, given a question $q_{new}$ and relevant paragraphs $\mathcal{P}_{rel}$, if the model's judgment aligns with whether the paragraphs are indeed related to the question, the judgment of$\mathcal{M}_{rel}$ is considered correct. For $\mathcal{M}_{decom}$, if both the LLM and annotators believe that each sub-question remains semantically consistent with the original question, the decomposition is considered effective.
\begin{table}[tp]
 \tabcolsep=0.5cm
 \resizebox{\linewidth}{!}{
\begin{tabular}{@{}cccc@{}}
\toprule
 & $\mathcal{M}_{know}$ & $\mathcal{M}_{rel}$ & $\mathcal{M}_{decom}$ \\ \midrule
Human & - & 93.5 & 89.5 \\
GPT4.0 & 97.0 & 95.0 & 87.0 \\ \bottomrule
\end{tabular}}
\caption{Human and GPT4 evaluation on the three models in RA-ISF.
}
\label{tab:two_assessments}
\end{table}
\vspace{5pt}

{
\textbf{The sub-modules results demonstrate high reliability.} The results are shown in Table~\ref{tab:two_assessments}, indicating that both human annotators and the large model consistently agree on the effectiveness of these three models, with accuracy rates exceeding 85\%. Specifically, $\mathcal{M}_{know}$ achieves an impressive accuracy of 97\%, suggesting a high cognitive ability of the trained model in recognizing its own knowledge. Meanwhile, the accuracy of $\mathcal{M}_{decom}$ is slightly lower, as the task of question decomposition falls within the realm of generative tasks, where there may be multiple feasible decomposition solutions. Overall, the three sub-modules exhibit high reliability in their respective tasks.

}


\section{Conclusions}

In this paper, we introduce RA-ISF, a framework designed to enhance retrieval augmentation effects and improve performance in open-domain question answering. 
This approach effectively mitigates the hallucination issues that are commonly seen in traditional retrieval augmentation and question-answering tasks. 
Experimental results demonstrate RA-ISF's superior performance across various benchmarks, and ablation studies validate the effectiveness of sub-modules.
Future research directions include further alleviating hallucination issues and improving the efficiency of the framework.

\section*{Limitation}

RA-ISF innovatively introduces a three-stage iterative problem-solving strategy. However, it's important to recognize its limitations and drawbacks. Firstly, iterative problem-solving can lead to an excessive branching of issues. In particular cases, this approach might become inefficient if it continuously explores a problem and its sub-problems without finding solutions or relevant passages. Secondly, different formulations of a problem may affect the effectiveness of the problem decomposition module, leading to small differences between the number of iterations and the outcome. 

Moreover, our method mainly relies on open-domain question-answering datasets. It has not been tested in specific fields such as mathematics reasoning, symbolic reasoning, or specialized areas like medicine and law. Future research could explore how it performs with these datasets. We also plan to investigate ways to use retrieval augmentation techniques more effectively and to simplify their complexity.

\section*{Ethics Statement}

Our approach employs the corpus of Wikipedia and utilizes open-source datasets for training and evaluating the model. All data are openly accessible. We leverage APIs for GPT-3.5 and open-source code and weights for Llama. Due to the hallucination issue of large language models, some of the generated content may contain factual errors and reasoning errors. RA-ISF offers a potential solution based on retrieval augmentation to mitigate the hallucination problem. Our work strictly adheres to the license and policies of
 released LLMs and publicly available datasets.

\bibliography{anthology,custom}
\bibliographystyle{acl_natbib}

\appendix

\label{sec:appendix}
\section{Details of Data Collection}\label{sec:appendix A}
\subsection{Data Collection of $\mathcal{M}_{know}$}
First, we use a QA dataset $\mathcal{D}$ for training, which includes questions $q_{i}$ and their corresponding correct answers $a_{i}$, represented as $\left \{ {q_{i},a_{i}} \right \}  _{i=1}^{\left |\mathcal{D}   \right | }$ . Initially, we extract the questions $q_{i}$ to query the LLM $\mathcal{M}$. Through few-shot prompts and in-context learning, we enable model $\mathcal{M}$ to generate answers for each question. In this scenario, the answers generated by the model rely entirely on internal knowledge.

We compare the model-generated answer $a_{g}$ with the correct answers $a_{i}$, and then categorize the questions $q_{i}$ into two groups. If $a_{g}$ is the same as $a_{i}$, then these questions fall into $\mathcal{Q}_{know}$, the category of problems that the model can solve on its own. Otherwise, these questions belong to $\mathcal{Q}_{unknow}$, the category of problems that the model cannot solve on its own. The specific expression is as follows:
\begin{equation}
q_{i}\in \begin{cases}  \mathcal{Q}_{know}& \text{ if } a_{i}=a_{g}  \\  \mathcal{Q}_{unknow}& \text{ if } a_{i}\ne a_{g}\end{cases}
\end{equation}

We collect various types of supervised training data and combine them to form the model's training data, ultimately resulting in the trained dataset $\mathcal{D}^{*} = \left \{ {\mathcal{Q}_{know},\mathcal{Q}_{unknow}} \right \}$.The $\mathcal{Q}_{know}$ class comprises questions that the model $\mathcal{M}$ inherently knows, while the $\mathcal{Q}_{unknow}$ class includes questions that the model is not aware of and requires external knowledge to obtain answers.

\subsection{Data Collection of $\mathcal{M}_{rel}$}
For a given $\mathcal{Q}$, we input it into the retriever $\mathcal{R}$, retrieving $k$ relevant paragraphs for each question $\mathcal{P}=$$\{P_{1},P_{2},...,P_{k}\}$. Subsequently, for each paragraph $P_{i}(i=1,2,...,k)$, we traverse them one by one, querying the LLM model $\mathcal{M}$ about the relevance of the retrieved paragraph $P_{i}$ to question $\mathcal{Q}$, and recording the model $\mathcal{M}$'s answer $\mathcal{A}=\{A_{1},A_{2},...,A_{k}\}$ where $\mathcal{A}=$ \texttt{relevant/irrelevant} for each paragraph.

We collect various types of supervised training data and combine them to form the model's training data, ultimately resulting in the trained dataset $\mathcal{D}^{*} = \left \{ {\mathcal{Q}+\mathcal{P},\mathcal{A}} \right \}$.

\subsection{Data Collection of $\mathcal{M}_{decom}$}
For a given $\mathcal{Q}$, we input it into the large model $\mathcal{M}$, instructing it to decompose each question utilized the prompt \texttt{"Please break down this question into several sub-questions and list them"}. For a given question $\mathcal{Q}$, the model breaks it down into k sub-questions, where the value of $k$ depends on the specific question. Finally, we document the sub-questions decomposed by the model for the question, denoted as $\mathcal{Q}_{sub}=\{q_{1},q_{2},...,q_{k}\}$.

We collect various types of supervised training data and combine them to form the model's training data, ultimately resulting in the trained dataset $\mathcal{D}^{*} = \left \{ {\mathcal{Q},\mathcal{Q}_{sub}} \right \}$.

\section{Details of Human Annotators}

The human annotators in Section \ref{sec:human} we invite all possess undergraduate or graduate degrees. We employ surveys, each containing a series of questions, to assess the results generated by the model. We inquire of the participants in the survey to provide their opinions on the relevance of the generated results to the questions and the correctness of the decomposition.

 \begin{figure}[tp]
    \centering
    \includegraphics[width=1\linewidth]{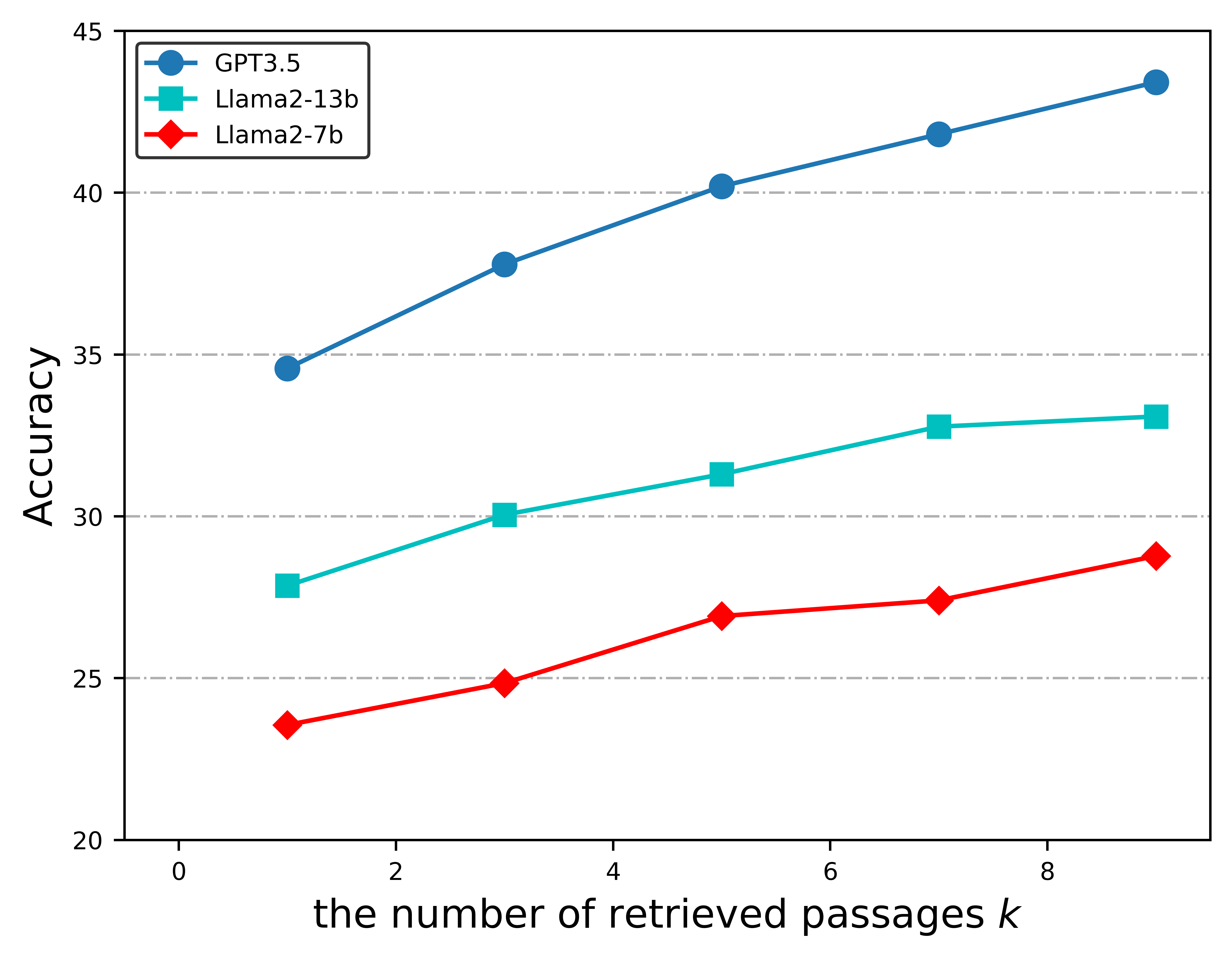}
    \caption{Trend of question accuracy on the NQ dataset with the growth of the iteration in question decomposition $k$.}
    \label{fig:k1-label}
\end{figure}

\section{Details of Datasets}
 \textbf{Natural Question (NQ)} \citep{kwiatkowski2019natural} is a question-answering dataset containing 307,373 training examples, 7,830 development examples, and 7,842 test examples. Each example is comprised of a google.com query and a corresponding Wikipedia page. 
 
\textbf{TriviaQA} \citep{joshi2017triviaqa} is a realistic text-based question-answering dataset that includes 950K question-answer pairs from 662K documents collected from Wikipedia and the web. For TriviaQA, given questions often have multiple valid answers, some of which are unsuitable for training targets, such as emoticons or spelling variations. Following \citet{lewis2020retrieval}, for TriviaQA, if a candidate answer does not appear in the top 1000 documents retrieved by the query, we filter it out.

\textbf{StrategyQA} \citep{geva2021did} is a question-answering benchmark where the required reasoning steps are implicit in the question, and should be inferred using a strategy. It includes 2,780 examples, each consisting of a strategy question, its decomposition, and evidence paragraphs. Questions in StrategyQA are short, topic-diverse, and cover a wide range of strategies.

 \textbf{HotpotQA} \citep{yang2018hotpotqa} is a multi-hop datasets from Wikipedia. The questions are diverse and not constrained to any pre-existing knowledge bases or knowledge schemas.
 HotpotQA is a question-answering dataset collected on the English Wikipedia, containing about 113K crowd-sourced questions that are constructed to require the introduction paragraphs of two Wikipedia articles to answer. Each question in the dataset comes with two gold paragraphs, as well as a list of sentences in these paragraphs that crowd workers identify as supporting facts necessary to answer the question.

 \textbf{2WikiMQA} \citep{ho2020constructing} utilizes both structured and unstructured data. In this dataset, evidence information is introduced, which includes reasoning paths for multi-hop questions. The evidence information serves two purposes: (i) providing a comprehensive explanation for predictions and (ii) evaluating the reasoning skills of a model. We carefully designed a pipeline and a set of templates during the generation of question-answer pairs to ensure the quality of multi-hop steps and questions.
 
 \begin{figure}[tp]
    \centering
    \includegraphics[width=1\linewidth]{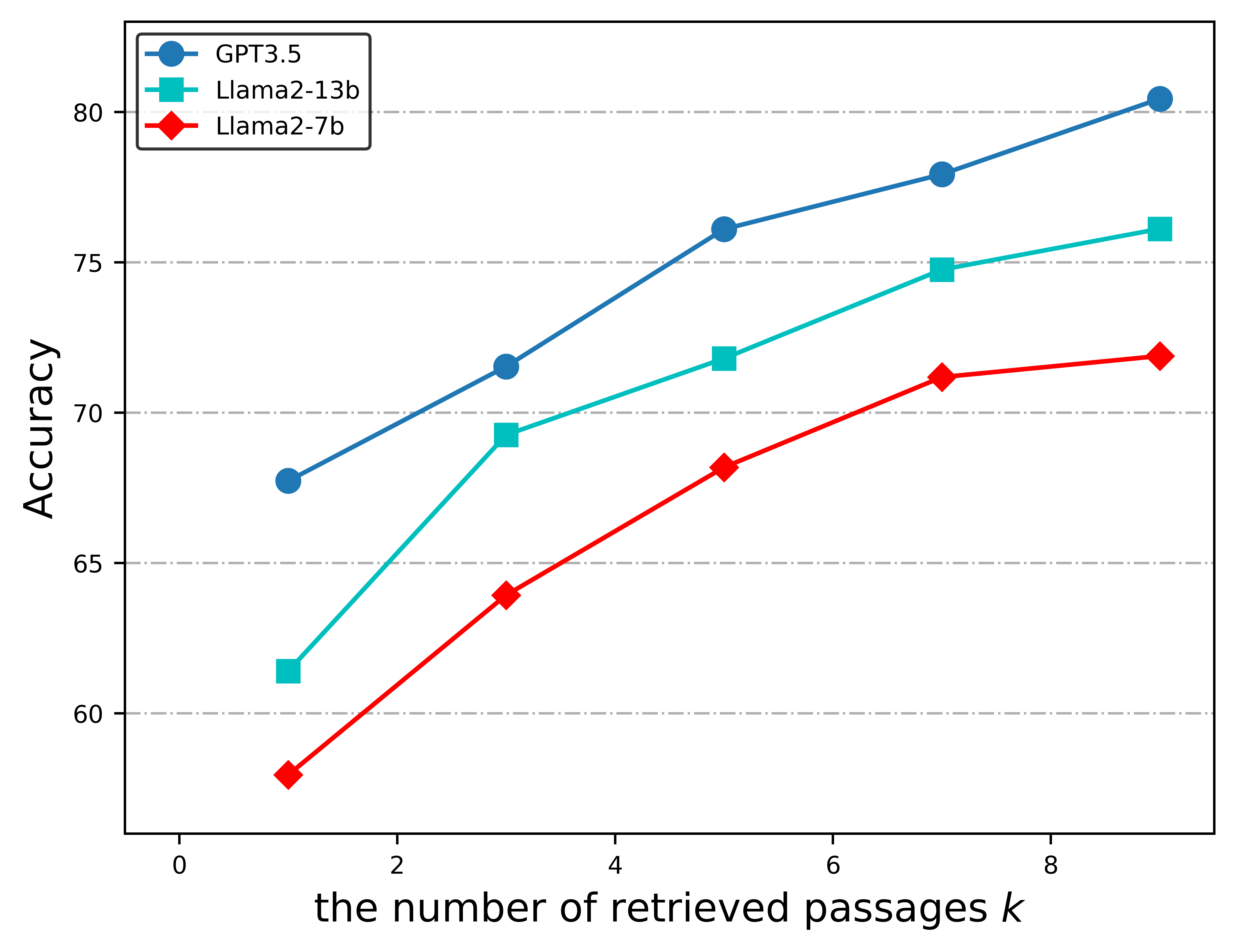}
    \caption{Trend of question accuracy on the NQ and TriviaQA dataset with the growth of the iteration in question decomposition $k$.}
    \label{fig:k2-label}
\end{figure}

\begin{table}[]
\resizebox{\linewidth}{!}{%
\begin{tabular}{@{}ccccc@{}}
\toprule
 \textbf{NQ} & \textbf{TriviaQA} & \textbf{HotpotQA} & \textbf{StrategyQA} & \textbf{2WikiMHQA} \\ \midrule
7830 & 17210 & 7405 & 1780 & 2935 \\ \bottomrule
\end{tabular} }
\caption{The number of datas of different datasets}
\label{tab:qa_datasets}
\end{table}

The number of datas in the test set in our experiments is shown as Table \ref{tab:qa_datasets}.

\begin{table*}[]
\resizebox{\linewidth}{!}{%
\begin{tabular}{lcccccccccc}
\toprule
\multicolumn{1}{c}{\multirow{2}{*}{}} & \multicolumn{2}{c}{\textbf{NQ}} & \multicolumn{2}{c}{\textbf{TriviaQA}} & \multicolumn{2}{c}{\textbf{HotpotQA}} & \multicolumn{2}{c}{\textbf{StrategyQA}} & \multicolumn{2}{c}{\textbf{2WikiMultihopQA}} \\ \cmidrule(lr){2-3} \cmidrule(lr){4-5} \cmidrule(lr){6-7} \cmidrule(lr){8-9} \cmidrule(lr){10-11} 
\multicolumn{1}{c}{}                  & WR         & Corr      & WR            & Corr         & WR            & Corr         & WR             & Corr          & WR               & Corr             \\ \midrule
Self-Knowledge                        & 65.8       & 2018      & 71.1          & 9586         & 24.3          & 761          & 52.1           & 732           & 31.5             & 308              \\
Passage-Relevant                      & 21.5       & 705       & 16.5          & 1809         & 32.8          & 1098         & 24.7           & 337           & 38.3             & 373              \\
Task-Decomposition                    & 12.7       & 425       & 12.4          & 1702         & 42.9          & 1584         & 23.2           & 282           & 30.2             & 379              \\ \bottomrule
\end{tabular}
}
\caption{The workload rates(WR) and correct problems solved(Corr) of different modules across various datasets.}
\label{tab:workload}
\end{table*}

\section{Analysis on the Number of Retrieved Passages}

 When the model is unable to solve a problem based solely on its own knowledge, we need to use a retriever to search for $k$ passages. In this regard, we need to investigate the values of $k$. Here, we experimented with NQ and TriviaQA datasets on models including GPT-3.5, $\text{Llama2}_\text{7B,13B}$,  with values of $k$ set to 1, 3, 5, 7, 9. The accuracy of the questions varies with the changes in $k$, as shown in Figure \ref{fig:k1-label} and \ref{fig:k2-label}.

\vspace{5pt}

{
\setlength{\parindent}{0cm}

\textbf{Increasing the number of retrieved passages helps improve the accuracy of problem-solving.} In general, as $k$ increases, the accuracy of the model in answering questions continues to improve. This is because increasing the number of retrieved paragraphs helps the model find more auxiliary knowledge, enhancing the likelihood of identifying relevant articles to the question and thereby improving the accuracy of question answering.

}

Further observation reveals that there is a noticeable improvement in accuracy as $k$ increases from 1 to 5; however, the improvement becomes less apparent when $k$ increases from 5 to 9. This is because, with the increase in the number of retrieved paragraphs, the model seems to have access to more paragraphs to assist in answering questions. However, in reality, the previously retrieved articles might have been sufficient for the model to identify the correct answers. Continuing to increase the number of retrievals could result in finding irrelevant articles, which would eventually be filtered out by $\mathcal{M}_{rel}$. Therefore, the contribution to the accuracy of the question is limited.

\section{Distribution of Problems Solving}
In this section, we aim to analyze the performance of different modules. Additionally, we investigate the distribution of solved problems among the three submodules. The experiments are conducted with GPT-3.5 as the base model and Llama2\textsubscript{7B} as the submodel.

To better observe the workload handled by each model, we use percentages as evaluation metrics.Simultaneously, we also evaluated the number of correct results in the test data to analyze the effectiveness and capabilities of different sub-modules. The result is shown on Table~\ref{tab:workload}.

\textbf{Different sub-modules have a certain workload.} For relatively simple questions, most are answered and resolved by the self-knowledge module. However, for complex datasets like 2wikimhqa, our framework tends to decompose complex questions into simpler sub-questions before addressing each part. This demonstrates our framework's capability in handling various types of problems.



\section{Integrating LLMs within Submodules}
Since the three submodules all gather responses from ChatGPT, which are subsequently utilized for training Llama, we conducted the following experiment to explore the impact of directly utilizing ChatGPT within these three submodules.

We adopt the distillation approach to help the small model learn the corresponding paradigm and save computational costs. Using ChatGPT directly could result in high API costs. As shown on Table~\ref{tab:submodel_size}, we found a limited improvement in performance. At this stage, we believe that the effectiveness of the responses primarily relies on the base model. Although better sub-models may offer some assistance, their impact is not significant.

\section{Examples}

\clearpage

\onecolumn
\begin{longtable}{@{}p{16cm}@{}}
\toprule
\multicolumn{1}{c}{\textbf{NQ}} \\
\toprule 
\textbf{Question}: When was the immigration reform and control act passed?\\
\textbf{Answer}: ["November 6, 1986" ] \\

\multicolumn{1}{c}{\textbf{Iteration 1}} \\
\textbf{$\mathcal{M}_{know}$}

\textbf{Instruction}: Can you use your own knowledge base to solve this problem? Answer yes if you know, no if you need additional knowledge base to solve it.

The problem is:When was the immigration reform and control act passed?

\textbf{Output}: Yes.

\textbf{Instruction}: Give the answer to the question.

\textbf{Final answer}: The Immigration Reform and Control Act (IRCA) was passed on November 6, 1986.\\ \bottomrule
\caption{An example from the NQ dataset.}
\label{tab:my-table4}
\end{longtable}

\begin{longtable}{@{}p{16cm}@{}}
\toprule
\multicolumn{1}{c}{\textbf{TriviaQA}} \\
\toprule 
\textbf{Question}: From which country did Angola achieve independence in 1975?\\
\textbf{Answer}: ["Portogało","Republic of Portugal"] \\

\multicolumn{1}{c}{\textbf{Iteration 1}} \\
\textbf{$\mathcal{M}_{know}$}

\textbf{Instruction}: Can you use your own knowledge base to solve this problem? Answer yes if you know, no if you need additional knowledge base to solve it.

The problem is: From which country did Angola achieve independence in 1975?

\textbf{Output}: Yes.

\textbf{Instruction}: Give the answer to the question.

\textbf{Final answer}: Angola achieved independence from Portugal in 1975.
\\ \bottomrule
\caption{An example from the TriviaQA dataset.}
\label{tab:my-table3}
\end{longtable}

\begin{longtable}{@{}p{16cm}@{}}
\toprule
\multicolumn{1}{c}{\textbf{HotpotQA}} \\
\toprule 
\textbf{Question}: Which private research university is located in Chestnut Hill, Massachusetts Boston College or Stanford University?\\
\textbf{Answer}: ["Boston College"] \\

\multicolumn{1}{c}{\textbf{Iteration 1}} \\
\textbf{$\mathcal{M}_{know}$}

\textbf{Instruction}: Can you use your own knowledge base to solve this problem? Answer yes if you know, no if you need additional knowledge base to solve it.

The problem is: Which private research university is located in Chestnut Hill, Massachusetts Boston College or Stanford University?

\textbf{Output}: Yes.

\textbf{Instruction}: Give the answer to the question.

\textbf{Final answer}: Boston College is the private research university located in Chestnut Hill, Massachusetts.\\ \bottomrule
\caption{An example from the HotpotQA.}
\label{tab:my-table5}
\end{longtable}

\onecolumn
\begin{longtable}{@{}p{16cm}@{}}
\toprule
\multicolumn{1}{c}{\textbf{2WikiMultihopQA}} \\
\toprule 
\textbf{Question}: Who is the mother of the director of film Polish-Russian War (Film)? \tabularnewline
\textbf{Answer}: ["Magorzata Braunek"] \tabularnewline
\multicolumn{1}{c}{\textbf{Iteration 1}} \tabularnewline
\textbf{$\mathcal{M}_{know}$} \tabularnewline
\textbf{Instruction}: Can you use your own knowledge base to solve this problem? Answer yes if you know, no if you need additional knowledge base to solve it. \tabularnewline
The problem is: Who is the mother of the director of film Polish-Russian War (Film)? \tabularnewline
\textbf{Output}: No \tabularnewline
\textbf{$\mathcal{M}_{rel}$} \tabularnewline
(Use retriever to find 5 passages...) \tabularnewline
\textbf{Instruction}: I will provide you with five additional pieces of knowledge based on the search for this question. Please assess whether these five paragraphs are relevant to the question and sufficient to answer it. If they are, please tell me what the relevant paragraphs are; if not, please answer "No." \tabularnewline

[1] Polish-Russian War (Wojna polsko-ruska) is a 2009 Polish film directed by Xawery Żuławski based on the novel Polish-Russian War under the white-red flag by Dorota Masłowska... \tabularnewline

[2] Xawery Żuławski (born 22 December 1971 in Warsaw) is a Polish film director. In 1995 he graduated National Film School in Łódź. He is the son of actress Małgorzata Braunek and director Andrzej Żuławski... \tabularnewline

[3] After an argument in a bar owned by "Left" (Michał Czernecki) "Strong" meets a "Gothgirl" Angelica (Maria Strzelecka) at night, an aspiring poet dressed in black, also a virgin and pessimist, for whom "suicide is a piece of cake"... \tabularnewline

[4] "Strong" follows Magda. He turns up at the town festival, where she takes part in a miss competition. He cannot reach her, but instead he meets... \tabularnewline

[5] Production The film was shot between May 6 and 18 June 2008 in locations of Warsaw, Wejherowo, Sopot and Gdynia outskirts. The film premiered on... \tabularnewline

\textbf{Output}: Relevant paragraphs: 
[2] Xawery Żuławski is the director of the film "Polish-Russian War (Wojna polsko-ruska)" and is the son of actress Małgorzata Braunek and director Andrzej Żuławski. \tabularnewline
\textbf{Instruction}: Using the knowledge from the relevant paragraphs, give the answer to the question. \tabularnewline
\textbf{Final answer}: The mother of the director of the film "Polish-Russian War (Wojna polsko-ruska)" is actress Małgorzata Braunek. \tabularnewline
\bottomrule
\caption{An example from the 2WikiMultihopQA dataset.}
\label{tab:my-table6}
\end{longtable}

\clearpage
\begin{longtable}{@{}p{16cm}@{}}
\toprule
\multicolumn{1}{c}{\textbf{StrategyQA}} \\
\toprule 
\textbf{Question}: Did Snoop Dogg refuse to make music with rival gang members?\\
\textbf{Answer}: ["false"]  \\

\multicolumn{1}{c}{\textbf{Iteration 1}} \\
\textbf{$\mathcal{M}_{know}$}

\textbf{Instruction}: Can you use your own knowledge base to solve this problem? Answer yes if you know, no if you need additional knowledge base to solve it.

The problem is: Did Snoop Dogg refuse to make music with rival gang members?

\textbf{Output}: No

\textbf{$\mathcal{M}_{rel}$}\\
(Use retriever to find 5 passages...)\\
\textbf{Instruction}: 
 I will provide you with five additional pieces of knowledge based on the search for this question. Please assess whether these five paragraphs are relevant to the question and sufficient to answer it. If they are, please tell me what the relevant paragraphs are; if not, please answer "No." 
 
[1] Calvin Cordozar Broadus Jr. ; born October 20, 1971), known professionally as Snoop Dogg (previously Snoop Doggy Dogg and briefly Snoop Lion)...

[2] Broadus' debut studio album, Doggystyle (1993), produced by Dr. Dre, was released by Death Row Records and debuted at number one on the Billboard 200...

[3] In 1993, Broadus was charged with first-degree murder for the shooting of a member of a rival gang who was actually killed by Snoop's bodyguard.Broadus was...

[4] While recording Doggystyle in August 1993, Broadus was arrested and charged with first-degree murder in connection with the shooting death of Philip Woldermariam...

[5] In 2002, he released the album Paid tha Cost to Be da Bo, on Priority/Capitol/EMI, selling over 1,310,000 copies. The album featured the hit singles...

\textbf{Output}: No.

\textbf{$\mathcal{M}_{decom}$}

\textbf{Instruction}: Please break down this question into several sub-questions and list them.

\textbf{Output}:

1. What is the name of the gang that Snoop Dogg is part of?

2. Which gang is the rival of the gang Snoop Dogg is associated with?

3. In Snoop Dogg's song "California Vacation," which rapper did he collaborate with?
\\
\multicolumn{1}{c}{\textbf{Iteration 2}} \tabularnewline

(Use sub-models to answer these sub-questions) \\

\textbf{Sub-answer}:

1.  Answer: American rapper Snoop Dogg is a member of the Crips gang.

2. Answer: The Crips are enemies of their rival gang, The Bloods.

3. Answer: Snoop Dogg collaborates with Xzibit and The Game on the song "California Vacation."

\textbf{Instruction}: Based on the sub-question answer, give the answer to the original question.

\textbf{Final answer}: No, Snoop Dogg did not refuse to make music with rival gang members, as evidenced by his collaboration with The Game on the song "California Vacation."\\ \bottomrule
\caption{An example from the StrategyQA dataset.}
\label{tab:my-table2}
\end{longtable}

\onecolumn
\begin{longtable}{@{}p{16cm}@{}}
\toprule
\multicolumn{1}{c}{\textbf{A Muti-iteration Example}}\\
\midrule
\textbf{Question}: Would Kelly Clarkson's voice shake glass?\\
\textbf{Answer}: ["true"] \\

\multicolumn{1}{c}{\textbf{Iteration 1}} \\
$\mathcal{M}_{know}$ \\

\textbf{Instruction}: Can you use your own knowledge base to solve this problem? Answer yes if you know, no if you need additional knowledge base to solve it. \\
The problem is: Would Kelly Clarkson's voice shake glass? \\
\textbf{Output}: No \\
$\mathcal{M}_{rel}$ \\
(Use retriever to find 5 passages...) \\
\textbf{Instruction}: I will provide you with five additional pieces of knowledge based on the search for this question. Please assess whether these five paragraphs are relevant to the question and sufficient to answer it. If they are, please tell me what the relevant paragraphs are; if not, please answer “No.” \\

[1] What exactly happened to Kelly Clarkson’s voice over the years?I understand…

[2] Back on what was at the time known as the “Farewell Season” of *American Idol*, Kelly Clarkson stunned…

[3] But it was that rendition of “O Holy Night” that had Kelly feeling a bit nervous before taking the stage...

[4] This season introduced a new albeit highly recognizable name looking for long-term syndicated success: original “American Idol” winner Kelly Clarkson...

[5] Kelly Clarkson (12,221 views):Significant High Notes:F$\sharp$6:(You Make Me Feel Like) A Natural Woman"live American Idol 2002);C$\sharp$6("Bad Romance"live The Kelly Clarkson Show… \\
\textbf{Output}: No. \\
$\mathcal{M}_{decom}$

\textbf{Instruction}: Please break down this question into several sub-questions and list them. \\
Certainly, breaking down the question "Would Kelly Clarkson's voice shake glass?" into sub-questions: \\
\textbf{Output}:\\
    1.{What factors determine whether a sound can break glass?}

   2.{What is the resonance frequency of glass?}

   3.{Can a human voice produce sound waves at frequencies capable of resonating with glass?}
   
   4.{Does Kelly Clarkson possess vocal abilities capable of generating sound waves at frequencies that could potentially resonate with glass?}
\\
\multicolumn{1}{c}{\textbf{Iteration2}} \\

\textbf{sub1:} \\
$\mathcal{M}_{know}$ \\
\textbf{Instruction}: Can you use your own knowledge base to solve this problem? Answer yes if you know, no if you need additional knowledge base to solve it. The problem is: What factors determine whether a sound can break glass? \\
\textbf{Output}: Yes. \\
\textbf{Instruction}: Give the answer to the question. \\
\textbf{Output}: The frequency of the sound wave needs to match the resonant frequency of the glass. If the sound wave resonates with the glass, it can lead to its fracture. \\
\midrule
\textbf{sub2:} \\
$\mathcal{M}_{know}$ \\
\textbf{Instruction}: Can you use your own knowledge base to solve this problem? Answer yes if you know, no if you need additional knowledge base to solve it. The problem is: What is the resonance frequency of glass? \\
\textbf{Output}: No. \\
\\
$\mathcal{M}_{rel}$\\
(Use retriever to find 5 passages...) \\
Instruction: I will provide you with five additional pieces of knowledge based on the search for this question. Please assess whether these five paragraphs are relevant to the question and sufficient to answer it. If they are, please tell me what the relevant paragraphs are; if not, please answer “No.” \\
\textbf{[1]} The glass are low at frequencies from 20-200 Hz and… \\
\textbf{[2]} The resonant frequency of wine glasses is typically within the range of human hearing (20-20,000 Hz)… \\
\textbf{[3]} Thus, for sound to break glass, it first has to match the resonant frequency of the said glass… \\
\textbf{[4]} Experiments of how sound breaks glass have shown that a minimum intensity of 105 dB is required… \\
\textbf{[5]} For glass to shatter, the sound must match the glass's natural resonant frequency (around 556 hertz)… \\
\textbf{Output}: The relevant paragraphs are: [1][2][3][4][5]. \\
\textbf{Instruction}: Using the knowledge from the relevant paragraphs, give the answer to the question. \\
\textbf{Output}: The resonance frequency of glass, particularly wine glasses, typically falls within the range of human hearing, which is between 20-20,000 Hz. However, for sound to break glass, it needs to match the glass's natural resonant frequency, which is around 400 Hz according to [1], or around 556 Hz according to [5]. Additionally, the sound must be sufficiently loud, with a minimum intensity of around 105 decibels, as stated in [4] and [5]. \\
\midrule
\textbf{sub3:} \\
$\mathcal{M}_{know}$\\
\textbf{Instruction}: Can you use your own knowledge base to solve this problem? Answer yes if you know, no if you need additional knowledge base to solve it. The problem is: Can a human voice produce sound waves at frequencies capable of resonating with glass? \\
\textbf{Output}: Yes. \\
\textbf{Instruction}: Give the answer to the question. \\
\textbf{Output}: Yes, a human voice can produce sound waves at frequencies capable of resonating with glass. This phenomenon is known as resonance. When the frequency of the sound waves produced by the human voice matches the natural frequency of the glass object, it can cause the glass to vibrate sympathetically, leading to resonance. This resonance can potentially lead to the glass shattering if the intensity of the sound waves is sufficient. \\
\midrule
\textbf{sub4:} \\
$\mathcal{M}_{know}$ \\
\textbf{Instruction}: Can you use your own knowledge base to solve this problem? Answer yes if you know, no if you need additional knowledge base to solve it. The problem is: Does Kelly Clarkson possess vocal abilities capable of generating sound waves at frequencies that could potentially resonate with glass? \\
\textbf{Output}: No. \\

$\mathcal{M}_{rel}$ \\
(Use retriever to find 5 passages...) \\
\textbf{Instruction}: I will provide you with five additional pieces of knowledge based on the search for this question. Please assess whether these five paragraphs are relevant to the question and sufficient to answer it. If they are, please tell me what the relevant paragraphs are; if not, please answer “No.” \\
\textbf{[1]} Kelly Clarkson's voice is just as strong as EVER! And she's even better vocally than she was 21 years ago… \\
\textbf{[2]} How can two sounds of the same frequency still sound different, for example sound of two different persons… \\
\textbf{[3]} If you need proof of Kelly Clarkson’s vocal prowess, just turn on your TV on a weekday afternoon… \\
\textbf{[4]} Peak expiratory flow (PEF) is a dynamic parameter and therefore requires a measuring device… \\
\textbf{[5]} And I look forward to sobbing along with Kelly’s new music as if I myself am a divorced mum-of-two – but… \\
\textbf{Output}: No. \\
\\
$\mathcal{M}_{decom}$ \\
\textbf{Instruction}: Please break down this question into several sub-questions and list them. \\
\textbf{Output}: \\
1. \textbf{What evidence exists regarding Kelly Clarkson's vocal abilities?} \\
2. \textbf{Is there any mention of the range or versatility of Kelly Clarkson's vocal abilities?} \\
3. \textbf{At what note would glass start to vibrate?} \\

\multicolumn{1}{c}{\textbf{Iteration 3}} \\
\textbf{sub4.1:} \\
$\mathcal{M}_{know}$ \\
\textbf{Instruction}: Can you use your own knowledge base to solve this problem? Answer yes if you know, no if you need additional knowledge base to solve it. \\
The problem is: What evidence exists regarding Kelly Clarkson's vocal abilities? \\
\textbf{Output}: No \\

$\mathcal{M}_{rel}$ \\
(Use retriever to find 5 passages...) \\
\textbf{Instruction}: I will provide you with five additional pieces of knowledge based on the search for this question. Please assess whether these five paragraphs are relevant to the question and sufficient to answer it. If they are, please tell me what the relevant paragraphs are; if not, please answer “No.” \\
\textbf{[1]} Kelly Clarkson's vocal range is approximately three octaves and a minor third, spanning Eb3 – C6 – F$\sharp$6. \\
\textbf{[2]} All of her songs have a certain of difficulty to sing. Also, she has been famous… \\
\textbf{[3]} \textbf{Vocal Range and Profile: Kelly Clarkson} \\
Range: C3 - C6 - G6 (F7) \\
Longest Note: 11 Seconds… \\
\textbf{[4]} Kelly Clarkson, a pop sensation and vocal virtuoso, has made an indelible mark on the music landscape… \\
\textbf{[5]} Simply put, it represents the distance between the highest and lowest tones a singer can reach, including… \\
\textbf{Output}: Relevant paragraphs: [1], [3], [4] \\
Not relevant: [2], [5] \\
\textbf{Instruction}: Using the knowledge from the relevant paragraphs, give the answer to the question. \\
\textbf{Output}: There is an indication that Kelly Clarkson's voice might have the capability to resonate with glass, as her vocal range is reported to span from Eb3 to F$\sharp$6, which covers a wide range of frequencies. \\
\midrule
\textbf{sub4.2:} \\
$\mathcal{M}_{know}$ \\
\textbf{Instruction}: Can you use your own knowledge base to solve this problem? Answer yes if you know, no if you need additional knowledge base to solve it. \\
The problem is: Is there any mention of the range or versatility of Kelly Clarkson's vocal abilities? \\
\textbf{Output}: Yes. Kelly Clarkson's vocal abilities are often praised for their range and versatility. She is known for her powerful voice, capable of reaching both high and low notes with ease, as well as for her ability to convey emotion and dynamics in her performances. Clarkson's vocal range spans several octaves, allowing her to tackle a variety of genres including pop, rock, country, and soul. Additionally, her versatility as a singer has been highlighted throughout her career, as she is able to adapt her voice to different styles and songs, showcasing her skill as a vocalist. \\
\midrule
\textbf{sub4.3:} \\
$\mathcal{M}_{know}$ \\
\textbf{Instruction}: Can you use your own knowledge base to solve this problem? Answer yes if you know, no if you need additional knowledge base to solve it. \\
The problem is: At what note would glass start to vibrate? \\
\textbf{Output}: Yes, I can provide an answer based on my current knowledge base. \\
The phenomenon you're referring to is known as resonance. When an object is subjected to vibrations at its natural frequency, it begins to resonate. For a glass, this frequency depends on its size, shape, and material properties. \\
Typically, wine glasses, for example, resonate at a frequency around the F$\sharp$ note above middle C (approximately 370 Hz), although this can vary depending on factors like the thickness of the glass and its shape. When you rub your finger along the rim of a wine glass, you can hear the glass vibrating at this frequency, producing a clear tone. If you were to match this frequency with a musical instrument, it would correspond to the F$\sharp$ note. \\
\midrule
\multicolumn{1}{c}{\textbf{Use the answer of sub 4.1-4.3 to answer sub4}} \\
\textbf{Instruction}: Base on the sub-question answer, give the answer to the origin question: \textbf{Does Kelly Clarkson possess vocal abilities capable of generating sound waves at frequencies that could potentially resonate with glass?} \\
\textbf{Output}: Yes, given that Kelly Clarkson's vocal range includes the F$\sharp$ note (F$\sharp$6), it is within the range of frequencies that could potentially resonate with glass. Therefore, based on the evidence provided, it is possible that Kelly Clarkson possesses vocal abilities capable of generating sound waves at frequencies that could potentially resonate with glass. \\
\midrule
\multicolumn{1}{c}{\textbf{Use the answer of sub1-4 to answer the question}} \\
\textbf{Instruction}: Base on the sub-question answer, give the answer to the origin question: \textbf{Would Kelly Clarkson's voice shake glass?} \\
\textbf{Output}: Based on the provided sub-question answers: \\
Yes, Kelly Clarkson's voice has the potential to shake glass. Given that her vocal abilities could generate sound waves within the range of frequencies capable of resonating with glass, if she were to produce sound waves at the resonant frequency of a particular glass object and with sufficient intensity, it could lead to the glass shattering. Therefore, Kelly Clarkson's voice has the capability to shake glass under the right conditions. \\
\bottomrule
\caption{A muti-iteration example.}
\end{longtable}

\end{document}